\documentclass[runningheads]{llncs}

\usepackage{eccv}

\usepackage{eccvabbrv}

\usepackage{graphicx}
\usepackage{booktabs}

\usepackage[accsupp]{axessibility}  
\usepackage{comment}

\usepackage[pagebackref,breaklinks,colorlinks,citecolor=eccvblue]{hyperref}

\usepackage{orcidlink}

\usepackage{graphicx}
\usepackage{amsmath}
\usepackage{amssymb}
\usepackage{booktabs}
\usepackage{multirow}
\usepackage{diagbox}
\usepackage{colortbl}
\usepackage{bm}
\usepackage{bbm}
\usepackage{mathrsfs}
\usepackage{amsmath}
\usepackage{threeparttable}
\usepackage[linesnumbered,ruled]{algorithm2e}
\DeclareMathOperator*{\argmax}{arg\,max}

\definecolor{lightgray}{gray}{0.92}
\definecolor{newgray}{rgb}{.6, .6, .6}

\begin{document}

\title{PartImageNet++ Dataset: Scaling up Part-based Models for Robust Recognition} 

\titlerunning{PartImageNet++ Dataset: Scaling up Part-based ...}

\author{Xiao Li\inst{1}\orcidlink{0000-0001-8992-4944} \and
Yining Liu\inst{2} \and
Na Dong\inst{3} \and
Sitian Qin\inst{2} \and
Xiaolin Hu\inst{1,4}\thanks{Corresponding author.}\orcidlink{0000-0002-4907-7354}
}

\authorrunning{Li et al.}

\institute{Department of Computer Science and Technology, BNRist, \\ Institute for Artificial Intelligence,
IDG/McGovern Institute for Brain Research, \\ Tsinghua Laboratory of Brain and Intelligence, Tsinghua University, Beijing, China \and
Harbin Institute of Technology, Weihai, China \and
Beijing Institute of Technology, Beijing, China \and
Chinese Institute for Brain Research (CIBR), Beijing, China \\
\email{lixiao20@mails.tsinghua.edu.cn, 22S030184@stu.hit.edu.cn, 3220220911@bit.edu.cn, qinsitian@hitwh.edu.cn, xlhu@mail.tsinghua.edu.cn}}

\maketitle

\begin{abstract}
Deep learning-based object recognition systems can be easily fooled by various adversarial perturbations. One reason for the weak robustness may be that they do not have part-based inductive bias like the human recognition process. Motivated by this, several part-based recognition models have been proposed to improve the adversarial robustness of recognition. However, due to the lack of part annotations, the effectiveness of these methods is only validated on small-scale nonstandard datasets. In this work, we propose PIN++, short for PartImageNet++, a dataset providing high-quality part segmentation annotations for all categories of ImageNet-1K (IN-1K). With these annotations, we build part-based methods directly on the standard IN-1K dataset for robust recognition. Different from previous two-stage part-based models, we propose a Multi-scale Part-supervised Model (MPM), to learn a robust representation with part annotations. Experiments show that MPM yielded better adversarial robustness on the large-scale IN-1K over strong baselines across various attack settings. Furthermore, MPM achieved improved robustness on common corruptions and several out-of-distribution datasets. The dataset, together with these results, enables and encourages researchers to explore the potential of part-based models in more real applications. The dataset and the code are available at \url{https://github.com/LixiaoTHU/PartImageNetPP}.

\end{abstract}

\section{Introduction}
\label{sec:intro}

Object recognition has achieved remarkable success due to the rise of Deep Neural Networks (DNNs) \cite{alexnet, resnet, convnext, swin}. However, DNNs often lack robustness and can be easily deceived by adversarial examples \cite{adv13, goodfellow14}, common image corruptions \cite{imagenetc}, and various out-of-distribution (OOD) shifts \cite{in-sketch, in-a-plus}, which greatly hinder applications of DNNs in security-critical scenarios \cite{realattack, advod}. One possible reason for the weakness may be that DNN-based recognition does not have a part-based inductive bias (\ie, recognition based on object parts) like the human recognition process, which is generally regarded as a robust system \cite{humanadv}. According to the well-known cognitive psychological theory recognition-by-components \cite{RBC}, humans prefer to recognize objects by decomposing objects into parts and taking into account the hierarchical representations and the spatial relationships of parts. This cognitive theory has been supported by numerous psychological evidence \cite{infants, opc, attentioneye, oldrbc}. However, several studies revealed that DNNs often exhibit a shortcut recognition behavior, \eg, relying on texture-based features \cite{sin} and even unintelligible features for humans \cite{bugfeature}.

\begin{figure}[!t]
  \centering
   \includegraphics[width=0.65\linewidth]{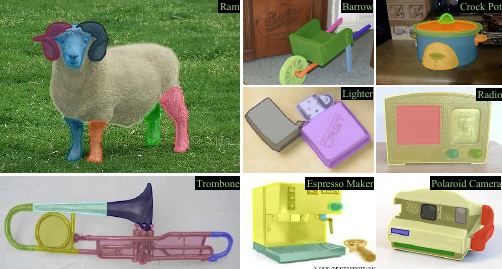}
   \caption{Examples of annotated images in PIN++. High-quality part segmentation annotations are provided on all categories in IN-1K. The object names are shown on the top-right of each image. The part names are hidden here.}
   \label{fig:dataset}
\end{figure}

If DNNs can perform a human-aligned part-based recognition, both the adversarial and non-adversarial robustness of DNNs could be closer to humans. Motivated by this idea, two recent works \cite{rock, carlinipart} investigated how to use additional part-level supervision to build up \textit{part-based (recognition) models}, which make predictions by recognizing the object's parts in a bottom-up manner. Both works investigated two-stage part-based models, where the first stage involves part segmentation, followed by the classification stage based on the segmentation results. These works have shown the potential of part-based models in enhancing robustness against various perturbations.

Nevertheless, due to the lack of part annotations, the effectiveness of part-based models has only been preliminary validated by the two works \cite{rock, carlinipart} on small-scale nonstandard datasets. The largest dataset for evaluation was PartImageNet (PIN) \cite{partimagenet}, consisting of about 24K images of 158 categories selected from the large-scale ImageNet-1K (IN-1K) \cite{imagenet}. PIN has several potential issues. Firstly, it is quite small compared to the standard IN-1K (about 1.3M images) commonly used in the domain of adversarial robustness \cite{transfer, bairobust}. This limited scale may restrict the generalizability of the findings. Secondly, categories included in PIN are specifically chosen to facilitate part-related tasks (\eg, part segmentation \cite{partseg} and object parsing \cite{objparsing}), with an emphasis on animal categories (118 out of 158). While this focus can be beneficial for studying part-related tasks themselves, it poses challenges in demonstrating the effectiveness of part-based models for general recognition tasks across a diverse range of objects. Thirdly, the annotation scheme of PIN is simple. It groups 158 categories into 11 super-categories and determines part annotations for each super-category, potentially failing to capture the characteristics of object categories within a super-category. Due to the three issues, the part-based models \cite{rock, carlinipart} trained on PIN cannot be directly compared with state-of-the-art (SOTA) methods trained on the standard IN-1K.

In this work, we introduce PartImageNet++ (PIN++), a new dataset providing high-quality part segmentation annotations for all categories of the widely used IN-1K \cite{imagenet}, to facilitate the research on part-based models for robust recognition (classification). To create PIN++, we designed a detailed annotation scheme to ensure high-quality part annotations. With this scheme, we annotated 100 randomly selected images per category, resulting in a total of 100K images with part annotations (about 1/13 of IN-1K). To avoid bias, all the part annotations are created manually without using any auxiliary model. To the best of our knowledge, PIN++ provides large-scale high-quality part segmentation annotations for the most diverse range of object categories, including creatures, artifacts, rigid objects, nonrigid objects, etc, among existing part datasets. \cref{fig:dataset} showcases some examples of annotated images. We introduce this dataset in detail in \cref{sec:dataset} and show that existing techniques fail to obtain high-quality part segmentation results without such annotations.

By leveraging the part annotations of PIN++, we develop part-based methods for robust recognition directly on the standard IN-1K dataset. To achieve this, we first train a part segmentation network with the annotations of PIN++ to obtain pseudo part labels for all unannotated images in IN-1K. We then propose a new Multi-scale Part-supervised recognition Model, termed MPM, to better exploit the part annotations. MPM is expected to learn a robust intermediate representation by adding auxiliary bypass layers to the vanilla recognition model (\eg, ResNet-50 \cite{resnet}). The auxiliary bypass layers are supervised by the part annotations and the generated pseudo part labels. Without introducing any additional parameter or computation during inference, MPM can use high-resolution part annotations more adequately than two-stage part-based models \cite{rock, carlinipart}, which are constrained to use low-resolution part labels. \cref{fig:structure} shows an overview of our whole method.

The experimental results demonstrate that MPM, combined with adversarial training (AT) \cite{at}, achieved better adversarial robustness on the large-scale IN-1K, outperforming strong AT baselines in both the seen and unseen adversarial threats. Furthermore, MPM exhibited improved robustness on common image corruptions \cite{imagenetc} and several OOD datasets. In addition, MPM achieved higher alignments with human vision. Beyond object recognition, MPM is shown to boost the adversarial robustness of downstream tasks such as object detection \cite{fasterrcnn}. The main contributions of this work can be summarized as follows:
\begin{itemize}
	\item We created PIN++, a new large-scale dataset with part annotations. This dataset can facilitate further research on part-based models for robust recognition and other part-related visual understanding tasks.
	
	\item We proposed MPM, a new part-supervised recognition model, to exploit the part annotations better, achieving superior robustness across various settings and benchmarks without introducing any extra inference cost.
\end{itemize}

\section{Related Work}
\label{sec:rw}

\noindent\textbf{Adversarial robustness.}
Szegedy \etal \cite{adv13} first revealed the vulnerability of DNNs to adversarial examples. Since then, numerous methods \cite{robustify, certifiedsm} have been proposed to enhance the adversarial robustness of DNNs. Among them, AT has emerged as the de facto paradigm for training adversarially robust DNNs. Early AT works \cite{trades, awp}  primarily focused on small-scale, low-resolution datasets such as CIFAR-10 \cite{cifar10}. Recently, there has been a growing interest in investigating AT on IN-1K \cite{imagenet}, as this dataset serves as a standard benchmark for evaluating SOTA computer vision techniques \cite{resnet, swin, convnext}. Several studies have demonstrated that AT models trained on IN-1K can transfer their adversarial robustness to downstream dense-prediction tasks and achieve zero-shot adversarially robust recognition \cite{advod, advseg, zeroshot}.

\noindent\textbf{Part datasets.}
The concept of \textit{parts} holds significant importance in both human visual cognition \cite{RBC} and computer vision domains. To conduct research on part-related visual tasks, the availability of part datasets is crucial. However, annotating object parts is challenging and expensive, and most part datasets are only limited to specific domains, such as \textit{cars} \cite{wang2015unsupervised, CarFusion, Apollocar3D} and \textit{humans} \cite{ATR,LIP,MHP,CIHP}. Furthermore, part datasets targeting common objects often have a limited number of object categories. Cityscapes PanopticParts (Cityscapes PP.) \cite{Cityscapes-Panoptic-Parts}, Pascal-Part \cite{PASCAL-Part}, and ADE20K provide part annotations for 5, 20, and 80 object categories, respectively. Although the recent PACO dataset \cite{paco} offers a large number of part annotations, it is still restricted to only 75 object categories. PIN \cite{partimagenet}, containing 158 categories from IN-1K \cite{imagenet}, represents the dataset with the most object categories in terms of part annotations.

\noindent\textbf{Part-based object recognition.}
Modeling objects in terms of parts has a long history in computer vision \cite{parthis, his1, his2}. But with the rise of DNNs, parts have been rarely used as auxiliaries for end-to-end general object recognition \cite{alexnet, resnet, swin, convnext}. Recently, two works \cite{rock, carlinipart} revisited part-based recognition models from the robustness perspective and showed the potential to enhance adversarial robustness. Li \etal \cite{rock} proposed ROCK, a part-based model that first predicts all parts represented by segmentation results, and then utilizes a judgment block to give category predictions based on the predicted parts. Similarly, Sitawarin \etal \cite{carlinipart} also investigated several types of two-stage part-based models, where the first stage involves part segmentation, followed by the recognition stage with a tiny classifier. However, as mentioned before, both works were only validated on small-scale part-friendly datasets for the lack of part annotations, \eg, Sitawarin \etal \cite{carlinipart} performed classification tasks only on 11 super-categories of PIN \cite{partimagenet}. Furthermore, the proposed two-stage part-based models inevitably introduced a lot of extra parameters during inference, while the utilization of high-resolution part annotations was limited, as discussed in \cref{sec:modeldesign}. In addition, ROCK \cite{rock} contains a non-differentiable process, which could lead to a potential overestimation of adversarial robustness \cite{AthalyeC018, adaptive20}. Our work will address all of these issues.

\section{PIN++ Dataset}
\label{sec:dataset}

We first present the details of how we built PIN++ to ensure high-quality part annotations, followed by the statistics of this dataset. We then give a discussion on PIN++.

\subsection{Annotation Scheme}
\label{sec:annotationscheme}

\textbf{Data source.} IN-1K \cite{imagenet}, containing about 1.3M images with category labels in the training set, is one of the most widely used datasets for general object recognition. Our aim is to provide part annotations for part-based robust recognition on IN-1K, and thus the images of PIN++ directly reuse the \textit{training} set of IN-1K. All the images conform to licensing for research purposes. However, due to cost constraints, it is impractical to provide part-level annotations for all training images in IN-1K. Instead, we decided to annotate 100 randomly selected images per category. Once part annotations are provided for all categories, modern supervised segmentation techniques \cite{maskrcnn} can be used to obtain pseudo part labels for the remaining unannotated images, as we will show in \cref{sec:method}. Note that PIN++ does not provide any part annotations for the validation set of IN-1K.

While PIN \cite{partimagenet} has provided some part annotations for 158 categories of IN-1K, their annotations do not fully meet our purpose and principles. To minimize annotation costs without compromising quality, we retained annotations of 90 categories from PIN as part of PIN++. The details are shown in \cref{sec:supp_reuse}. We then show the annotation design for the remaining 910 categories of IN-1K.

\noindent\textbf{Annotation quality control.} To ensure high-quality part annotations, three points should be taken into account:

\noindent\textbf{1) Deciding which parts to annotate per category.} One of the key challenges in the part annotation task is the ambiguity of object part selection (\eg, how to annotate the parts of \textit{hammer}). PIN \cite{partimagenet} evades this problem and focuses primarily on annotating parts for animal categories (most \textit{quadrupeds} can be regarded as consisting of four part categories: \textit{head}, \textit{body}, \textit{foot}, \textit{tail}). However, the remaining categories in IN-1K are highly diverse, making it challenging to decide which parts to annotate. To ensure a scientifically grounded division of parts for each category, we initially consult the Wikidata knowledge base\footnote{\url{https://www.wikidata.org/wiki/Wikidata:Introduction}} to obtain part vocabularies for every object category (\eg, \textit{hammer} consists of three parts according to Wikidata: \textit{handle}, \textit{striker}, and \textit{hammerhead}). Additionally, for object categories without a clear definition of parts on Wikidata, we decide which parts to annotate by asking recruited volunteers which parts they think to be important to their cognition of the object. For example, according to the cognition of volunteers, for most \textit{quadruped} categories, annotating \textit{head}, \textit{body}, \textit{foot}, and \textit{tail} is enough, while for \textit{ram} (see \cref{fig:dataset} for one example), \textit{horn} should be annotated as an extra part. For categories that are indeed difficult to decompose into parts, \eg, \textit{flatworm}, we generalize the concept of ``parts'' and treat the foreground of an object as one part category. In this case, the part is the object itself. Other details on deciding the parts to annotate are further discussed in \cref{sec:supp_part}. These principles ensure that all categories in IN-1K can have part annotations.

\noindent\textbf{2) Designing the part segmentation principles.} We design several part segmentation principles to guide the annotators and ensure high-quality part segmentation annotation. Firstly, the annotated part masks are required to be combined to cover the entire object and have no overlap, unless in the case of the second principle. Secondly, for some parts that indeed should overlap in semantic concepts (\eg, \textit{horn} and \textit{head}), the annotators should annotate the inclusion relation of parts (\eg, \textit{horn} is included in \textit{head}), and then these part masks can be annotated with overlap. Thirdly, to keep the consistency of part annotations among images of the same object category, following \cite{partimagenet}, the same annotator should annotate all 100 images of this object category. The other principles are shown in \cref{sec:supp_principle}.

\noindent\textbf{3) Annotation quality inspection.} Ten randomly selected images together with the annotation visualizations (see \cref{fig:dataset}) are sent to inspectors to assess if the annotations for a specific object category meet the defined principles. If the annotations for any two images fail to meet these principles, the entire category's annotations are re-annotated until the requirements are satisfied.

\begin{table}[!tb]
  \centering
  \caption{The numbers of annotated object categories, part categories, images and part masks of publicly available part datasets.}
  \small
  \setlength{\tabcolsep}{4pt}
  {
    \begin{tabular}{c|cccc}
     Dataset  & Object Category & Part Category & Image & Part Mask  \\
     \midrule
     Cityscapes PP. \cite{Cityscapes-Panoptic-Parts} & $5$ & $23$ & $3.5$K & $100$K \\
     Pascal-Part \cite{PASCAL-Part} & $20$ & $193$ & $19$K & $363.5$K \\
     ADE20K$^\star$ \cite{ade20k}  & $80$ & $566$ & $12.6$K & $193.2$K \\
     PACO \cite{paco} & $75$ & $456$ & $76.7$K& $\mathbf{641.4}$K \\
     PIN \cite{partimagenet} & $158$ & $609$ & $24$K & $112$K \\
     \rowcolor{lightgray} PIN++ & $\mathbf{1000}$ & $\mathbf{3308}$ & $\mathbf{100}$K & $406.4$K  \\
    \end{tabular}
    \begin{tablenotes}
            \footnotesize
            \item $^\star$: ADE20K is severely category-unbalanced and here we show statistics for categories with more than 10 annotated part masks.
    \end{tablenotes}
    }
  \label{tab:comparedataset}
\end{table}

\subsection{Statistics}

We finally annotated 100 images per category for 910 categories in IN-1K, resulting in a total of 91K
images with part annotations. Taking these data and the data of 90 reused categories of PIN together, PIN++ for 1K categories is obtained. Following IN-1K, here we ensure PIN++ is class-balanced. PIN++ is split into \texttt{train/val} sets with a 9:1 ratio for each category. But note that all annotated images are part of the training set of IN-1K. Unless stated otherwise, they are all utilized for training in our following experiments and the evaluation is performed on IN-1K \texttt{val} set.

In summary, PIN++ includes a total of 100K images with part annotations. These annotations cover 3,310 part categories for 1,000 object categories. A total of 406,364 part masks are annotated in PIN++. Further statistics of PIN++ are shown in \cref{sec:supp_statistics}.

\noindent\textbf{Comparison with other part datasets.} 
We compare PIN++ with publicly available part datasets, as shown in \cref{tab:comparedataset}. PIN++ outperforms previous datasets in terms of the number of object and part categories, as well as the total number of images with annotated part masks. It also demonstrates the competitive number of part masks compared to the recent PACO dataset \cite{paco}. Notably, PIN++ offers part annotations for a diverse range of objects, including creatures, artifacts, rigid objects, and nonrigid objects, unlike PACO or PIN, which are primarily focused on common tools or animals, respectively. As shown in \cref{sec:supp_statistics}, the annotation quality of PIN++ exhibits an advantage over that of PIN. Moreover, PIN++ is category-balanced and guarantees that each image contains only one foreground category. All these features build the unique advantage of PIN++ for studying object recognition tasks.

\subsection{Discussion on PIN++}
\label{sec:dispart}

\begin{figure}[!tb]
  \centering
   \includegraphics[width=0.85\linewidth]{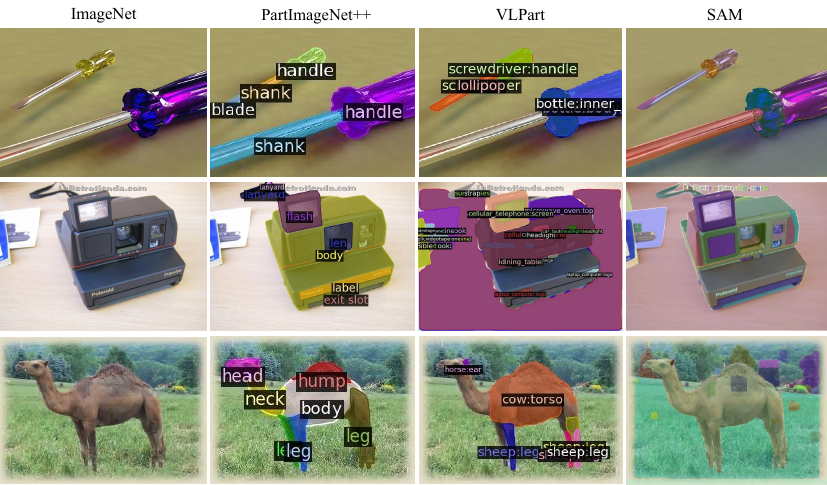}
   \caption{Comparison between part segmentation results of different methods and PIN++ annotations. Without training on PIN++, VLPart and SAM fail to segment objects into specific parts with accurate semantics.}
   \label{fig:datasetcompare}
\end{figure}

Recent advancements in segmentation include CLIP-based open-vocabulary segmentation methods \cite{glip, opendet, vlpart} and the SAM model \cite{sam}. We compared these techniques with the part annotations of PIN++. We used the largest version of SAM, and for open-vocabulary segmentation, we employed the VLPart model \cite{vlpart}, which is trained with part annotations from several part datasets including Pascal-Part \cite{PASCAL-Part}, PIN \cite{partimagenet}, and PACO \cite{paco}. Visual comparisons are presented in \cref{fig:datasetcompare}. We can see that VLPart only provides roughly reasonable semantics for part categories it has been trained on. It fails to accurately identify the semantics of parts from unseen objects. On the other hand, SAM seems to perform well in segmenting certain parts based on the object's edges, but it struggles when the part edges are less distinguishable, as shown in the segmentation results for \textit{camel} (the third row in \cref{fig:datasetcompare}). In addition, note that all masks segmented by SAM are category-agnostic.

The above results highlight the challenges faced by existing techniques in achieving accurate part segmentation for diverse objects without training on PIN++. Additionally, current text-to-image generative models \cite{stablediffusion} also struggle with understanding part semantics \cite{diffusion}. However, supervised training on PIN++ yields good part segmentation results, as shown in \cref{sec:supp_preudo}.
Therefore, besides its usage in robust object recognition (as demonstrated next), we believe that PIN++ and its extensive collections of 3308 part categories (vocabularies) have the potential to enhance general part-related visual understanding tasks.

\section{Part-based Recognition Method}
\label{sec:method}
\begin{figure*}[!t]
  \centering
   \includegraphics[width=\linewidth]{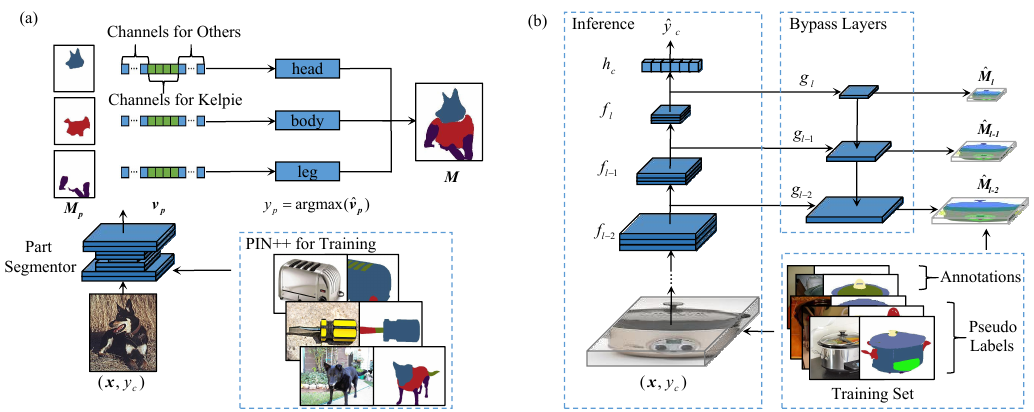}
   \caption{An overview of the generation of pseudo-labels and the structure of MPM. (a) A part segmentation model trained on PIN++ and used to obtain pseudo-part labels for unannotated images. (b) MPM adds several auxiliary bypass layers to the vanilla recognition model for part segmentation supervision. MPM is trained by part annotations together with the pseudo part labels. During inference, the auxiliary layers are dropped, and the vanilla recognition model gives the final object category prediction. }
   \label{fig:structure}
\end{figure*}

With the annotations of PIN++, we develop part-based methods directly on the standard IN-1K. Our part-based methods can be divided into two steps. We first obtain pseudo part labels for the remaining unlabeled images of IN-1K and then train the MPM with these part annotations and pseudo labels. The pipeline of our part-based methods is illustrated in \cref{fig:structure}.

\subsection{Pseudo Part Label Generation} 
\label{pseudo-generation}

\noindent\textbf{Part segmentation model.}
Arbitrary instance segmentation models \cite{maskrcnn, instanceasquery} can be leveraged to perform part segmentation and generate pseudo part labels when trained with the part annotations from PIN++. Following Sun \etal \cite{vlpart}, we use the classical Mask R-CNN \cite{maskrcnn} model with a vision transformer Swin-B \cite{swin} as the backbone network. In training the Mask R-CNN, we consider the masks for each part category as masks for individual object categories directly. During inference for obtaining pseudo part labels, we incorporate a simple post-processing operation alongside the regular inference procedure.

\noindent\textbf{Post-processing.}
Given an image $\mathbf{x} \in \mathbb{R}^{3 \times H \times W}$ with object category $y_c$, where $H \times W$ represents the resolution, the regular inference procedure of the trained Mask R-CNN produces pseudo part labels in the form of $\{(\mathbf{M_p}, \mathbf{v_p})\}$, where $\mathbf{M_p} \in \{0, 1\} ^ {H \times W}$ denotes the binary part mask, $\mathbf{v_p} \in [0, 1] ^ {K}$ denotes the output probability of part categories, and $K$ denotes the number of part categories (3308 in our case). But note that the images that need inference are all in the training set of IN-1K. Their object category $y_c$ is not utilized in the regular inference procedure. To leverage it, we apply a Category Filter (CF) operation to obtain $\mathbf{\hat{v}_p}$, which sets the probabilities of unrelated part categories in $\mathbf{v_p}$ to zero (\eg, for \textit{cat}, except for \textit{cat:head}, \textit{cat:body}, etc, other part categories such as \textit{radio:button} and \textit{bird:head} are all excluded). The pseudo part category for $\mathbf{M_p}$ is then computed as $y_p = \argmax \mathbf{\hat{v}_p}$.

\subsection{Multi-scale Part-supervised Recognition Model}
\label{sec:modeldesign}
\noindent\textbf{Data preparation.}
After obtaining the pseudo labels for all remaining training images of IN-1K, we simply treat the pseudo labels and real part annotations equally because we find that the quality of the pseudo labels is roughly satisfactory, as shown in \cref{sec:supp_preudo}. The part labels $\{(\mathbf{M_p}, \mathbf{y_p})\}$ for an image $\mathbf{x}$ are then converted to a single composite segmentation mask $\mathbf{M} \in \{0, 1\} ^ {H \times W \times (K + 1)}$, where the extra one channel indicates the background. Finally, the data samples for training part-based models are $\{(\mathbf{x}, y_c, \mathbf{M})\}$.

\noindent\textbf{Model design.} 
Both previous works \cite{rock, carlinipart} build part-based models in the two-stage way: a segmenter $\mathcal{F}_{\mathrm{seg}}: \mathbb{R} ^ {3 \times H \times W}\rightarrow\mathbb{R} ^ {(K+1) \times h \times w}$ and an extra module $\mathcal{F}_{\mathrm{cls}}: \mathbb{R} ^ {(K+1) \times h \times w}\rightarrow\mathbb{R}^{C}$ for classification, where $h \times w$ denotes the output resolution and $C$ denotes the number of object category. $\mathcal{F}_{\mathrm{seg}}$ is a backbone network (\eg, a ResNet-50 without classification head). Most popular backbones \cite{resnet, convnext, swin} are hierarchically composed of several blocks with down-sampling layers and thus they can be written as $\mathcal{F}_{\mathrm{seg}} := f_l \circ \cdots \circ f_1$, where $l$ denotes the number of down-sampling layers and $f_{i(1 \leq i \leq l)}$ denotes one block with the $i$th down-sampling layer. $\mathcal{F}_{\mathrm{cls}}$ can be a tiny classifier \cite{carlinipart} or a non-differentiable judgment block \cite{rock} and its input is the part segmentation results $\hat{\mathbf{M}} = \mathcal{F}_{\mathrm{seg}}(\mathbf{x})$. The overall model for recognition is  $\mathcal{F} := \mathcal{F}_{\mathrm{cls}}\circ\mathcal{F}_{\mathrm{seg}}$. However, as $K$ is significantly larger than $C$, using $\hat{M}$ as the intermediate results for recognition inevitably introduce extra parameters and computation during inference. This situation gets worse with an increased number of part categories, as shown in \cref{sec:ablation}. Moreover, these methods solely employ part supervision on the output of $f_l$. Without expanding $\mathcal{F}_{\mathrm{seg}}$, part annotations have to be down-sampled to match the output resolution of $f_l$, potentially limiting the utilization of high-resolution part annotations.

Instead of these two-stage part-based models, MPM is expected to learn a robust intermediate representation by adding multi-scale bypass layers to the vanilla recognition network. In MPM, the model used for object recognition is the vanilla backbone network directly: $\mathcal{F} := h_c \circ f_l \circ \cdots \circ f_1$, where $h_c$ denotes a vanilla classification head (usually a linear layer). MPM utilizes the part annotations to supervise the intermediate features of $\mathcal{F}$ by several bypass layers $g_i$: $\hat{\mathbf{M}_i} = g_i \circ f_i \circ \cdots \circ f_1(\mathbf{x})$, where $i \leq l$. In this way, the intermediate features of $\mathcal{F}$ can be seen as implicit representations of part segmentation results when $g_i$ is simple enough. Note that here we use part supervision not only on the output of $f_l$, but also $f_{i(i<l)}$. As the output of lower layers has a larger resolution, it can be supervised by part annotations with higher resolution. But the outputs of $f_{i(i<l)}$ are relatively low-level, which may not be enough to obtain high-level part segmentation results themselves. To balance this, only the outputs of the last three blocks $f_{i(l - 2 \leq i \leq l)}$ are supervised by part mask $\mathbf{M}$ (\eg, for a ResNet-50 with input size $224 \times 224$, the intermediate features of $7 \times 7$, $14 \times 14$, and $28 \times 28$ are supervised by corresponding down-sampled $\mathbf{M}$, respectively). In addition, following the principle of FPN \cite{fpn}, some top-down layers are used to augment the low-level features with high-level features. Note that distinct from the conventional FPN that generally has comparable parameters with the backbone network \cite{fpn}, these bypass layers are extremely lightweight as our goal is to improve the recognition accuracy rather than the quality of part segmentation results. \cref{fig:structure}(b) shows the overall structure of MPM. During inference, these auxiliary layers are dropped, and the vanilla recognition model gives the final object category prediction.

\noindent\textbf{Training objective.} The overall loss for training MPM is: $L = L_{\mathrm{cls}} + \lambda \cdot L_{\mathrm{seg}}$, where $L_{\mathrm{cls}}$ denotes the vanilla loss for classification, $L_{\mathrm{seg}}$ denotes the loss for part segmentation, and $\lambda$ is a hyper-parameter. We compute $L_{\mathrm{seg}}$ as the average of losses on three part segmentation results of different resolutions, as mentioned before. Note that when performing AT, adversarial examples are generated by $\mathbf{x}^{\star} = \argmax_{\mathbf{x}^{\star}: ||\mathbf{x}^{\star} - \mathbf{x}||_p \leq \epsilon} L_{\mathrm{cls}}(\mathcal{F}(\mathbf{x}^{\star}), y)$, while MPM is always trained with $L$.

\section{Experiments}
\label{sec:expr}

\subsection{Experimental Setup}
\label{sec:setting}

\textbf{Training setup.} Unless specified otherwise, following previous works \cite{bairobust, debenedetti}, we performed AT with $l_\infty$ bound $\epsilon = 4/255$ on IN-1K. The inner optimization for obtaining adversarial examples used PGD \cite{at} with iterative steps $t = 2$ and the step size was set to be $s = 2 * \epsilon / t$. The input resolution used $224 \times 224$. We used the vanilla version of the AT \cite{at} and did not incorporate other variants such as TRADES \cite{trades} or AWP \cite{awp}. These variants have shown effectiveness mainly on small-scale datasets, making their generalization to IN-1K non-trivial. Following previous part-based works \cite{rock, carlinipart}, we used the classical ResNet-50 \cite{resnet} network with 25.6M parameters as the baseline model. We note that recent works \cite{bairobust, debenedetti} used a GELU \cite{gelu} activation to replace the original RELU \cite{relu} activation of ResNet-50 for boosting robustness when performing AT, here we also used GELU, denoted as ResNet-50-Gelu. In addition, we improved the training recipe of ResNet-50 to build a strong baseline. Other details on the training recipe are provided in \cref{sec:supp_recipe}. Note that bypass layers (\eg, point-wise convolutions) with about 4.5M parameters were used by MPM during training but completely dropped during inference. The code is submitted along with the paper.

\noindent\textbf{Evaluation setup.}
For all experiments in the work, unless specified otherwise, we evaluated the adversarial robustness with AutoAttack \cite{autoattack}, which is a combination of various attack methods and is generally recognized as a reliable evaluation \cite{robustbench}. The evaluation was performed at a resolution $224 \times 224$ on randomly selected 10K images of the 50K IN-1K \texttt{val} set, 10 images for each category. This ensures that the 95\% confidence intervals of the reported average adversarial robustness (see \cref{tab:reeval}) were less than $\pm0.5\%$. Note that, unlike previous works \cite{bairobust, debenedetti}, we did not use the 5K images selected by RobustBench \cite{robustbench} as we found that they were category-unbalanced, containing three categories even without any image for validation.

We expect the AT models with $l_\infty$ bound $\epsilon = 4/255$ to be adversarially robust not only on seen threats (\ie, $l_\infty$ attack with $\epsilon = 4/255$) but also on unseen attack threats (\eg, $l_\infty$ attack with larger $\epsilon$, $l_1$ and $l_2$ attacks). Thus, we evaluated the models on three attack threats: $l_\infty$, $l_1$, and $l_2$, with the bounds $\epsilon_\infty = 4/255$, $\epsilon_1 = 75$, $\epsilon_2 = 2$, respectively. For $l_\infty$ attack, we extra evaluated the models with $\epsilon_\infty = 8/255$ to mimic a stronger threat, denoted as $l_\infty(\times 2)$.

\begin{table*}[!t]
  \centering
  \caption{Recognition accuracies ($\%$) of methods under different attack threats on IN-1K, following the evaluation setup in \cref{sec:setting}. \textit{Average} denotes the average accuracies in four attack settings. We highlight the best results in each column.}
  \small
   \setlength{\tabcolsep}{4pt}
   {
    \begin{tabular}{cc|ccccc|c}
	  \multirow{2}*{Architecture}	& \multirow{2}*{Method} & \multicolumn{6}{c}{Adversarial Train wrt $l_\infty$ ($\epsilon = 4/255$)} \\
	  	  &  &  Clean & $l_\infty$ & $l_\infty (\times 2)$ & $l_1$ & $l_2$ &  Average  \\
    \midrule
	  
	  ResNet-50 & \cite{transfer}, 2020 & $63.9$ & $35.9$ & $13.2$ & $2.0$ & $13.7$ & $16.2$ \\
	  ResNet-50 & \cite{easyrobust}, 2022 & $64.7$ & $34.3$ & $11.4$ & $2.3$ & $16.4$ & $16.1$ \\
        ResNet-50 & \cite{liu}, 2023 & $65.5$ & $32.6$ & $9.5$ & $3.3$ & $17.9$ & $15.8$ \\
        ResNet-50-Gelu & \cite{bairobust}, 2021 & $\mathbf{67.8}$ & $36.6$ & $10.8$ & $3.0$ & $16.8$ & $16.8$ \\
	  ResNet-50-Gelu & \cite{debenedetti}, 2023 & $66.8$ & $35.6$ & $12.8$ & $3.6$ & $16.5$ & $17.1$ \\
	  \rowcolor{lightgray} ResNet-50-Gelu & ours (vanilla) & $67.1$ & $38.1$ & $12.6$ & $5.0$ & $21.6$ & $19.3$ \\
	  \rowcolor{lightgray} ResNet-50-Gelu & ours (MPM) & $\mathbf{67.8}$ & $\mathbf{39.1}$ & $\mathbf{13.6}$ & $\mathbf{6.2}$ & $\mathbf{24.3}$ & $\mathbf{20.8}$ \\
	  
    \end{tabular}
    }
  \label{tab:reeval}%
\end{table*}

\subsection{Robustness against Adversarial Attacks}
\label{sec:results}
With the above setup, we trained the vanilla ResNet-50-Gelu as a baseline. MPM used the same recipe as the baseline model for a fair comparison, except that it used an extra $L_\mathrm{seg}$ to introduce part-based supervision. $\lambda$ was simply set to 1. In addition, we compared MPM with \textit{all} open-sourced ResNet-50 checkpoints adversarially trained on IN-1K in recent years. The recognition accuracies under different attack threats are shown in \cref{tab:reeval}. We can see that our baseline model is strong enough, outperforming all other benchmarks by 2.2\% on average robustness (17.1\% \vs 19.3\%). MPM further shows significantly better adversarial robustness than the vanilla baseline across all these attack threats. In addition, compared with our strong vanilla baseline, the accuracy of clean images is also improved with MPM. To our best knowledge, these are SOTA results of adversarial robustness with ResNet-50 on IN-1K. We additionally show in \cref{sec:supp_comparison} that these results are significantly better than previous results on PIN.

\begin{table*}[!t]
  \centering
  \caption{Recognition accuracies (\%) of ResNet-50-Gelu on 15 different common image corruptions \cite{imagenetc}. We adopt accuracy as the metric to be consistent with other results, while this metric can be easily converted into the corruption error metric \cite{imagenetc}. ``Average'' denotes the accuracy averaged over different common corruptions. }
  \small
  \resizebox{\linewidth}{!}{
  \setlength{\tabcolsep}{0.7pt}
  {
    \begin{tabular}{cc|ccc|cccc|cccc|cccc|c}
    \multirow{2}*{Part} & \multirow{2}*{AT} & \multicolumn{3}{c|}{Noise}   & \multicolumn{4}{c|}{Blur} & \multicolumn{4}{c|}{Weather}  & \multicolumn{4}{c|}{Digital}& \multirow{2}*{Avg.}\\
                                     &                          & Gauss.  & Shot & \multicolumn{1}{c|}{Impul.}  & Defoc.  & Glass & Moti. & \multicolumn{1}{c|}{Zoom}  & Snow & Frost & Fog & \multicolumn{1}{c|}{Bright}  & Cont.  & Elast.  & Pixel  & JPEG \\
    \midrule
    $ $ & $ $  &$31.2$&$30.7$&$27.3$&$39.0$&$28.6$&$39.5$&$\mathbf{37.8}$&$35.8$&$40.7$&$53.8$&$69.0$&$38.0$&$\mathbf{37.0}$&$47.1$&$57.0$&$40.8$ \\
    $\checkmark$ & $ $ &$\mathbf{31.5}$&$\mathbf{31.0}$&$\mathbf{28.4}$&$\mathbf{41.9}$&$\mathbf{29.8}$&$\mathbf{40.3}$&$36.7$&$\mathbf{37.0}$&$\mathbf{41.3}$&$\mathbf{55.3}$&$\mathbf{69.6}$&$\mathbf{39.6}$&$36.9$&$\mathbf{52.9}$&$\mathbf{59.9}$&$\mathbf{42.1}$ \\
    \hline
    $ $ & $\checkmark$ &$29.6$&$28.5$&$21.0$&$23.3$&$32.9$&$32.8$&$34.4$&$34.2$&$34.0$&$10.8$&$58.4$&$9.5$&$52.2$&$58.8$&$63.1$&$34.9$ \\
    $\checkmark$ & $\checkmark$ &$\mathbf{30.1}$&$\mathbf{29.2}$&$\mathbf{22.4}$&$\mathbf{25.2}$&$\mathbf{35.1}$&$\mathbf{35.0}$&$\mathbf{36.4}$&$\mathbf{36.4}$&$\mathbf{36.2}$&$\mathbf{11.9}$&$\mathbf{59.7}$&$\mathbf{10.4}$&$\mathbf{53.1}$&$\mathbf{59.8}$&$\mathbf{64.3}$&$\mathbf{36.4}$ \\

    \end{tabular}
     }
     }
  \label{tab:corruption}%
\end{table*}%

\begin{table}[!t]
  \centering
  \caption{Recognition accuracies (\%) of models on four OOD datasets.}
  \small
  \resizebox{0.6\linewidth}{!}{
 \setlength{\tabcolsep}{4pt}
  {
    \begin{tabular}{cc|cccc|c}
     Part & AT  & IN-A-Plus & IN-Sketch & SIN & DIN  & Average  \\
     \midrule
     $ $ & $ $ & $6.9$ & $\mathbf{25.8}$ & $6.9$ & $53.5$ & $23.3$\\
     $\checkmark$ & $ $ & $\mathbf{7.4}$ & $25.7$ & $\mathbf{7.4}$ & $\mathbf{53.9}$ & $\mathbf{23.6}$\\
     \hline
     $ $ & $\checkmark$ & $5.3$ & $25.1$ & $12.5$ & $55.4$ & $24.6$\\
     $\checkmark$ & $\checkmark$ & $\mathbf{5.7}$ & $\mathbf{26.4}$ & $\mathbf{12.6}$ & $\mathbf{55.9}$ & $\mathbf{25.2}$\\
    \end{tabular}   
    }
    }
  \label{tab:odd-dataset}
\end{table}

\subsection{Robustness on Corruptions and OOD Datasets}
\label{sec:common}
Except for robustness against adversarial attacks, which consider the worst-case security threat scenarios, we evaluated the robustness of our part-based model on common image corruptions \cite{imagenetc} and several OOD datasets \cite{in-a-plus, sin, in-sketch, modelvshuman}, which could be more realistic threats. Considering that AT can have negative effects on these non-adversarial scenarios \cite{rock, liu}, here we also evaluated models without AT. The training details are also shown in \cref{sec:supp_recipe}.

For common image corruptions \cite{imagenetc}, we generated different types of corrupted images on the whole IN-1K \texttt{val} set. The results of different models on these images are shown in \cref{tab:corruption}, where ``Part'' indicates MPM. Here the results of each corruption type are the accuracies averaged on five severity levels \cite{imagenetc}. It is seen that MPM has better robustness on nearly all types of image corruptions than models without part supervision, regardless of AT. Besides common corruptions, we evaluated the models on four OOD datasets: ImageNet-A-Plus (IN-A-Plus) \cite{in-a-plus}, ImageNet Sketch (IN-Sketch) \cite{in-sketch}, Stylized ImageNet (SIN) \cite{sin}, and \textit{image distortion} dataset (denoted as DIN) \cite{modelvshuman}. See \cref{sec:supp_datasets} for the introduction of these datasets. The results shown in \cref{tab:odd-dataset} indicate that MPM also exhibits improved robustness on these OOD shifts.

\subsection{Other Advantages of MPM}
\noindent\textbf{Models versus humans.} 
The part-based models are motivated by the human recognition process. We were interested in whether the robustness of MPM aligned more with human cognition. To investigate this, we employed the evaluation method introduced by Geirhos \etal \cite{modelvshuman}, which involved comparing the models' decisions on different distorted images with the actual judgments made by human observers. The results shown in \cref{sec:supp_humanvsmodel} indicate that MPM measurably improved alignment with the human recognition process.

\noindent\textbf{Boosting the downstream task.}
Li \etal \cite{advod} show that adversarially trained models on the large-scale IN-1K can be utilized to initialize the backbone of downstream networks (\eg, Faster R-CNN \cite{fasterrcnn}). This approach together with downstream AT enables the transfer of adversarial robustness to downstream tasks. We conducted similar experiments by using the checkpoints of the vanilla baseline and MPM. The results shown in \cref{sec:supp_downstream} indicate that MPM enhances both the clean accuracy and adversarial robustness in object detection.

\begin{table}[!t]
  \centering
  \caption{Segmentation accuracies (AP) of Mask R-CNN with and without CF, and recognition accuracies (\%) of MPM on IN-1K trained with various pseudo labels.}
  \small
  \setlength{\tabcolsep}{4pt}
  {
    \begin{tabular}{cc|cc|ccc}

     Pseudo Label & CF & AP & AP$_{50}$ & Clean & $l_\infty$  & $l_\infty (\times 2)$ \\
     \midrule
      $$ & $$  & - & - & $65.5$ & $33.0$ & $9.4$\\
       $\checkmark$ & $ $ &  $37.2$ & $58.6$ & $\mathbf{67.8}$ & $38.8$ & $13.1$\\
       $\checkmark$ & $\checkmark$  & $\mathbf{40.7}$ & $\mathbf{64.4}$ & $\mathbf{67.8}$ & $\mathbf{39.1}$ & $\mathbf{13.6}$\\
    \end{tabular}  
    }
  \label{tab:pl_quality}
\end{table}

\subsection{Ablation Study}
\label{sec:ablation}

\textbf{Pseudo-label quality.}
We first investigate the influence of pseudo-label quality on the adversarial robustness of MPM. In \cref{pseudo-generation}, we illustrate the use of CF during generating pseudo labels. Here we remove the CF operator and regenerate pseudo labels for images without part annotations. The left half of \cref{tab:pl_quality} (the 2nd and 3rd rows) shows the segmentation accuracies of models with and without CF (here the models were supervised with the \texttt{train} set of PIN++, and the AP were calculated on \texttt{val}). It is seen that training with PIN++ achieved good part segmentation results, indicating its potential for enhancing part-related tasks. CF further significantly improves AP.

We then trained MPM using both pseudo labels and real part annotations, as well as solely using real part annotations. The results, displayed in \cref{tab:pl_quality} (the 1st row), indicate that training MPM solely with real part annotations seems to be ineffective, potentially compromising the model's performance. We guess it could caused by overfitting on the images with part annotations. Besides, the inclusion of better-quality pseudo labels can improve the robustness of our part-based model, underscoring the importance of high-quality pseudo labels.

\noindent\textbf{Part-based model structure.} 
We then conducted ablation experiments on the structure of the part-based models. Specifically, we compared MPM with the two-stage part-based model designed by Sitawarin \etal \cite{carlinipart}. In addition, we performed ablations on the number of part supervisions. MPM incorporates part supervision with three resolutions, referred to as SV3. Here we compared it with SV1 and SV2, which indicate part-based models solely supervised by $f_l$ and both $f_l$ and $f_{l-1}$, respectively. Besides these, we extra performed an ablation by removing all top-down (TD) connections of MPM. The results are shown in \cref{tab:diff_model_structure}. We can see that although the two-stage model needs about twice the parameters during inference, its robustness is inferior to the MPM. And all the designs of MPM, including utilizing higher resolution part annotations and top-down connections, contribute to the robustness.

\begin{table}[!t]
  \centering
  \caption{Recognition accuracies (\%) of different part-based models on IN-1K. The number of parameters during inference is listed.}
  \small
 \setlength{\tabcolsep}{4pt}
  {
    \begin{tabular}{c|c|ccc}
     Method & Param. &  Clean &$l_\infty$ & $l_\infty (\times 2)$  \\
     \midrule
     Two-stage & $60.9$M &  $67.3$ & $38.7$ & $13.0$\\
     \hline
     SV1 & \multirow{4}*{$25.6$M} &  $67.3$ & $36.9$ & $8.8$\\
     SV2 w/ TD &   & $67.6$ & $37.1$ & $11.7$\\
     MPM (SV3 w/ TD)&  &  $\mathbf{67.8}$ & $\mathbf{39.1}$ & $\mathbf{13.6}$\\
     SV3 w/o TD & & $67.4$ & $36.9$ & $11.7$ \\
     
    \end{tabular}   
    }
  \label{tab:diff_model_structure}
\end{table}

\noindent\textbf{Additional ablations.}
Additional ablation experiments emphasize the significance of fine-grained part annotations rather than object segmentation annotations. Experiments on Tiny-ImageNet \cite{tinyimagenet} highlight the significance of investigating adversarial robustness on high-resolution images. Moreover, the adversarial robustness of part-based models exhibits low sensitivity to $\lambda$. Refer to \cref{sec:supp_others} for comprehensive information about these ablation studies.

\section{Conclusion and Discussion}
\label{sec:conclusion}
In this work, we introduce PIN++, a new large-scale high-quality part dataset, to facilitate the research on part-based models for robust recognition. We propose a new MPM to leverage this dataset. With extensive experiments, we demonstrate the robustness of MPM across diverse scenarios. But we note that MPM represents just a stepping stone in the realm of part-based models for robustness. We hope that this work will inspire and motivate researchers to further explore the potential of part-based models for robust recognition.

\noindent\textbf{Limitation.} 
We believe that except for robust recognition, PIN++ has the potential to boost general part-related visual understanding tasks. We have not conducted many benchmark experiments specifically for such tasks because they are beyond the scope of this work. We leave these as future work. 


\section*{Acknowledgement}
This work was supported in part by the National Key Research and Development Program of China (No. 2021ZD0200301) and the National Natural Science Foundation of China (No. U2341228).

%
%
\bibliographystyle{splncs04}
\bibliography{main}

\begin{thebibliography}{10}
\providecommand{\url}[1]{\texttt{#1}}
\providecommand{\urlprefix}{URL }
\providecommand{\doi}[1]{https://doi.org/#1}

\bibitem{relu}
Agarap, A.F.: Deep learning using rectified linear units (relu). arXiv preprint
  arXiv:1803.08375  (2018)

\bibitem{AthalyeC018}
Athalye, A., Carlini, N., Wagner, D.A.: Obfuscated gradients give a false sense
  of security: Circumventing defenses to adversarial examples. In: Int. Conf.
  Mach. Learn. (ICML). vol.~80, pp. 274--283 (2018)

\bibitem{bairobust}
Bai, Y., Mei, J., Yuille, A.L., Xie, C.: Are transformers more robust than
  cnns? In: Adv. Neural Inform. Process. Syst. (NeurIPS). pp. 26831--26843
  (2021)

\bibitem{RBC}
Biederman, I.: Recognition-by-components: a theory of human image
  understanding. Psychological Review  \textbf{94}(2), ~115 (1987)

\bibitem{his1}
Burl, M.C., Weber, M., Perona, P.: A probabilistic approach to object
  recognition using local photometry and global geometry. In: Eur. Conf.
  Comput. Vis. (ECCV). vol.~1407, pp. 628--641 (1998)

\bibitem{PASCAL-Part}
Chen, X., Mottaghi, R., Liu, X., Fidler, S., Urtasun, R., Yuille, A.L.: Detect
  what you can: Detecting and representing objects using holistic models and
  body parts. In: IEEE Conf. Comput. Vis. Pattern Recog. (CVPR). pp. 1979--1986
  (2014)

\bibitem{robustbench}
Croce, F., Andriushchenko, M., Sehwag, V., Debenedetti, E., Flammarion, N.,
  Chiang, M., Mittal, P., Hein, M.: Robustbench: a standardized adversarial
  robustness benchmark. In: NeurIPS Datasets and Benchmarks (2021)

\bibitem{autoattack}
Croce, F., Hein, M.: Reliable evaluation of adversarial robustness with an
  ensemble of diverse parameter-free attacks. In: Int. Conf. Mach. Learn.
  (ICML). Proceedings of Machine Learning Research, vol.~119, pp. 2206--2216
  (2020)

\bibitem{advseg}
Croce, F., Singh, N.D., Hein, M.: Robust semantic segmentation: Strong
  adversarial attacks and fast training of robust models. Adv. Neural Inform.
  Process. Syst. (NeurIPS)  (2023)

\bibitem{debenedetti}
Debenedetti, E., Sehwag, V., Mittal, P.: A light recipe to train robust vision
  transformers. In: First IEEE Conference on Secure and Trustworthy Machine
  Learning (SaTML). pp. 225--253 (2023)

\bibitem{imagenet}
Deng, J., Dong, W., Socher, R., Li, L., Li, K., Fei{-}Fei, L.: Imagenet: {A}
  large-scale hierarchical image database. In: IEEE Conf. Comput. Vis. Pattern
  Recog. (CVPR). pp. 248--255 (2009)

\bibitem{realattack}
Eykholt, K., Evtimov, I., Fernandes, E., Li, B., Rahmati, A., Xiao, C.,
  Prakash, A., Kohno, T., Song, D.: Robust physical-world attacks on deep
  learning visual classification. In: IEEE Conf. Comput. Vis. Pattern Recog.
  (CVPR). pp. 1625--1634 (2018)

\bibitem{instanceasquery}
Fang, Y., Yang, S., Wang, X., Li, Y., Fang, C., Shan, Y., Feng, B., Liu, W.:
  Instances as queries. In: Int. Conf. Comput. Vis. (ICCV). pp. 6890--6899
  (2021)

\bibitem{parthis}
Fischler, M.A., Elschlager, R.A.: The representation and matching of pictorial
  structures. {IEEE} Trans. Comput.  \textbf{22}(1),  67--92 (1973)

\bibitem{modelvshuman}
Geirhos, R., Narayanappa, K., Mitzkus, B., Thieringer, T., Bethge, M.,
  Wichmann, F.A., Brendel, W.: Partial success in closing the gap between human
  and machine vision. In: Adv. Neural Inform. Process. Syst. (NeurIPS). pp.
  23885--23899 (2021)

\bibitem{sin}
Geirhos, R., Rubisch, P., Michaelis, C., Bethge, M., Wichmann, F.A., Brendel,
  W.: Imagenet-trained cnns are biased towards texture; increasing shape bias
  improves accuracy and robustness. In: Int. Conf. Learn. Rep. (ICLR) (2019)

\bibitem{his2}
Girshick, R.B., Iandola, F.N., Darrell, T., Malik, J.: Deformable part models
  are convolutional neural networks. In: IEEE Conf. Comput. Vis. Pattern Recog.
  (CVPR). pp. 437--446 (2015)

\bibitem{CIHP}
Gong, K., Liang, X., Li, Y., Chen, Y., Yang, M., Lin, L.: Instance-level human
  parsing via part grouping network. In: Eur. Conf. Comput. Vis. (ECCV). vol.
  11208, pp. 805--822 (2018)

\bibitem{goodfellow14}
Goodfellow, I.J., Shlens, J., Szegedy, C.: Explaining and harnessing
  adversarial examples. In: Int. Conf. Learn. Rep. (ICLR) (2015)

\bibitem{generated}
Gowal, S., Rebuffi, S., Wiles, O., Stimberg, F., Calian, D.A., Mann, T.A.:
  Improving robustness using generated data. In: Adv. Neural Inform. Process.
  Syst. (NeurIPS). pp. 4218--4233 (2021)

\bibitem{infants}
Haaf, R.A., Fulkerson, A.L., Jablonski, B.J., Hupp, J.M., Shull, S.S.,
  Pescara-Kovach, L.: Object recognition and attention to object components by
  preschool children and 4-month-old infants. Journal of Experimental Child
  Psychology  \textbf{86}(2),  108--123 (2003)

\bibitem{partimagenet}
He, J., Yang, S., Yang, S., Kortylewski, A., Yuan, X., Chen, J., Liu, S., Yang,
  C., Yu, Q., Yuille, A.L.: Partimagenet: {A} large, high-quality dataset of
  parts. In: Eur. Conf. Comput. Vis. (ECCV). vol. 13668, pp. 128--145 (2022)

\bibitem{maskrcnn}
He, K., Gkioxari, G., Doll{\'{a}}r, P., Girshick, R.B.: Mask {R-CNN}. IEEE
  Trans. Pattern Anal. Mach. Intell. (TPAMI)  \textbf{42}(2),  386--397 (2020)

\bibitem{resnet}
He, K., Zhang, X., Ren, S., Sun, J.: Deep residual learning for image
  recognition. In: IEEE Conf. Comput. Vis. Pattern Recog. (CVPR). pp. 770--778
  (2016)

\bibitem{imagenetc}
Hendrycks, D., Dietterich, T.G.: Benchmarking neural network robustness to
  common corruptions and perturbations. In: Int. Conf. Learn. Rep. (ICLR)
  (2019)

\bibitem{gelu}
Hendrycks, D., Gimpel, K.: Gaussian error linear units (gelus). arXiv preprint
  arXiv:1606.08415  (2016)

\bibitem{naturaladv}
Hendrycks, D., Zhao, K., Basart, S., Steinhardt, J., Song, D.: Natural
  adversarial examples. In: IEEE Conf. Comput. Vis. Pattern Recog. (CVPR). pp.
  15262--15271 (2021)

\bibitem{attentioneye}
Hoffman, J.E., Subramaniam, B.: The role of visual attention in saccadic eye
  movements. Perception \& Psychophysics  \textbf{57}(6),  787--795 (1995)

\bibitem{bugfeature}
Ilyas, A., Santurkar, S., Tsipras, D., Engstrom, L., Tran, B., Madry, A.:
  Adversarial examples are not bugs, they are features. In: Adv. Neural Inform.
  Process. Syst. (NeurIPS). pp. 125--136 (2019)

\bibitem{ema}
Izmailov, P., Podoprikhin, D., Garipov, T., Vetrov, D.P., Wilson, A.G.:
  Averaging weights leads to wider optima and better generalization. In: {UAI}.
  pp. 876--885 (2018)

\bibitem{sam}
Kirillov, A., Mintun, E., Ravi, N., Mao, H., Rolland, C., Gustafson, L., Xiao,
  T., Whitehead, S., Berg, A.C., Lo, W.Y., Dollar, P., Girshick, R.: Segment
  anything. In: Int. Conf. Comput. Vis. (ICCV). pp. 4015--4026 (2023)

\bibitem{cifar10}
Krizhevsky, A., Hinton, G., et~al.: Learning multiple layers of features from
  tiny images  (2009)

\bibitem{alexnet}
Krizhevsky, A., Sutskever, I., Hinton, G.E.: Imagenet classification with deep
  convolutional neural networks. In: Adv. Neural Inform. Process. Syst.
  (NeurIPS). pp. 1106--1114 (2012)

\bibitem{tinyimagenet}
Le, Y., Yang, X.: Tiny imagenet visual recognition challenge. CS 231N
  \textbf{7}(7), ~3 (2015)

\bibitem{glip}
Li, L.H., Zhang, P., Zhang, H., Yang, J., Li, C., Zhong, Y., Wang, L., Yuan,
  L., Zhang, L., Hwang, J., Chang, K., Gao, J.: Grounded language-image
  pre-training. In: IEEE Conf. Comput. Vis. Pattern Recog. (CVPR). pp.
  10955--10965 (2022)

\bibitem{advod}
Li, X., Chen, H., Hu, X.: On the importance of backbone to the adversarial
  robustness of object detectors. arXiv preprint arXiv:2305.17438  (2023)

\bibitem{in-a-plus}
Li, X., Li, J., Dai, T., Shi, J., Zhu, J., Hu, X.: Rethinking natural
  adversarial examples for classification models. arXiv preprint
  arXiv:2102.11731  (2021)

\bibitem{rock}
Li, X., Wang, Z., Zhang, B., Sun, F., Hu, X.: Recognizing object by components
  with human prior knowledge enhances adversarial robustness of deep neural
  networks. IEEE Trans. Pattern Anal. Mach. Intell. (TPAMI)  \textbf{45}(7),
  8861--8873 (2023)

\bibitem{zeroshot}
Li, X., Zhang, W., Liu, Y., Hu, Z., Zhang, B., Hu, X.: Language-driven anchors
  for zero-shot adversarial robustness. In: IEEE Conf. Comput. Vis. Pattern
  Recog. (CVPR) (2024)

\bibitem{LIP}
Liang, X., Gong, K., Shen, X., Lin, L.: Look into person: Joint body parsing
  {\&} pose estimation network and a new benchmark. IEEE Trans. Pattern Anal.
  Mach. Intell. (TPAMI)  \textbf{41}(4),  871--885 (2019)

\bibitem{ATR}
Liang, X., Liu, S., Shen, X., Yang, J., Liu, L., Dong, J., Lin, L., Yan, S.:
  Deep human parsing with active template regression. IEEE Trans. Pattern Anal.
  Mach. Intell. (TPAMI)  \textbf{37}(12),  2402--2414 (2015)

\bibitem{opendet}
Lin, C., Sun, P., Jiang, Y., Luo, P., Qu, L., Haffari, G., Yuan, Z., Cai, J.:
  Learning object-language alignments for open-vocabulary object detection. In:
  Int. Conf. Learn. Rep. (ICLR) (2023)

\bibitem{fpn}
Lin, T., Doll{\'{a}}r, P., Girshick, R.B., He, K., Hariharan, B., Belongie,
  S.J.: Feature pyramid networks for object detection. In: IEEE Conf. Comput.
  Vis. Pattern Recog. (CVPR). pp. 936--944 (2017)

\bibitem{focal}
Lin, T., Goyal, P., Girshick, R.B., He, K., Doll{\'{a}}r, P.: Focal loss for
  dense object detection. IEEE Trans. Pattern Anal. Mach. Intell. (TPAMI)
  \textbf{42}(2),  318--327 (2020)

\bibitem{coco}
Lin, T.Y., Maire, M., Belongie, S., Hays, J., Perona, P., Ramanan, D.,
  Doll{\'a}r, P., Zitnick, C.L.: Microsoft coco: Common objects in context. In:
  Eur. Conf. Comput. Vis. (ECCV). pp. 740--755 (2014)

\bibitem{liu}
Liu, C., Dong, Y., Xiang, W., Yang, X., Su, H., Zhu, J., Chen, Y., He, Y., Xue,
  H., Zheng, S.: A comprehensive study on robustness of image classification
  models: Benchmarking and rethinking. arXiv preprint arXiv:2302.14301  (2023)

\bibitem{diffusion}
Liu, Q., Kortylewski, A., Bai, Y., Bai, S., Yuille, A.L.: Intriguing properties
  of text-guided diffusion models. arXiv preprint arXiv:2306.00974  (2023)

\bibitem{partseg}
Liu, S., Zhang, L., Yang, X., Su, H., Zhu, J.: Unsupervised part segmentation
  through disentangling appearance and shape. In: IEEE Conf. Comput. Vis.
  Pattern Recog. (CVPR). pp. 8355--8364 (2021)

\bibitem{swin}
Liu, Z., Lin, Y., Cao, Y., Hu, H., Wei, Y., Zhang, Z., Lin, S., Guo, B.: Swin
  transformer: Hierarchical vision transformer using shifted windows. In: Int.
  Conf. Comput. Vis. (ICCV). pp. 9992--10002 (2021)

\bibitem{convnext}
Liu, Z., Mao, H., Wu, C., Feichtenhofer, C., Darrell, T., Xie, S.: A convnet
  for the 2020s. In: IEEE Conf. Comput. Vis. Pattern Recog. (CVPR). pp.
  11966--11976 (2022)

\bibitem{at}
Madry, A., Makelov, A., Schmidt, L., Tsipras, D., Vladu, A.: Towards deep
  learning models resistant to adversarial attacks. In: Int. Conf. Learn. Rep.
  (ICLR) (2018)

\bibitem{easyrobust}
Mao, X., Chen, Y., Li, X., Qi, G., Duan, R., Zhang, R., Xue, H.: Easyrobust: A
  comprehensive and easy-to-use toolkit for robust computer vision.
  \url{https://github.com/alibaba/easyrobust} (2022)

\bibitem{Cityscapes-Panoptic-Parts}
Meletis, P., Wen, X., Lu, C., de~Geus, D., Dubbelman, G.:
  Cityscapes-panoptic-parts and pascal-panoptic-parts datasets for scene
  understanding. arXiv preprint arXiv:2004.07944  (2020)

\bibitem{paco}
Ramanathan, V., Kalia, A., Petrovic, V., Wen, Y., Zheng, B., Guo, B., Wang, R.,
  Marquez, A., Kovvuri, R., Kadian, A., Mousavi, A., Song, Y., Dubey, A.,
  Mahajan, D.: {PACO:} parts and attributes of common objects. In: IEEE Conf.
  Comput. Vis. Pattern Recog. (CVPR). pp. 7141--7151 (2023)

\bibitem{CarFusion}
Reddy, N.D., Vo, M., Narasimhan, S.G.: Carfusion: Combining point tracking and
  part detection for dynamic 3d reconstruction of vehicles. In: IEEE Conf.
  Comput. Vis. Pattern Recog. (CVPR). pp. 1906--1915 (2018)

\bibitem{fasterrcnn}
Ren, S., He, K., Girshick, R.B., Sun, J.: Faster {R-CNN:} towards real-time
  object detection with region proposal networks. In: Adv. Neural Inform.
  Process. Syst. (NeurIPS). pp. 91--99 (2015)

\bibitem{stablediffusion}
Rombach, R., Blattmann, A., Lorenz, D., Esser, P., Ommer, B.: High-resolution
  image synthesis with latent diffusion models. In: IEEE Conf. Comput. Vis.
  Pattern Recog. (CVPR). pp. 10674--10685 (2022)

\bibitem{oldrbc}
Rosch, E., Mervis, C.B.: Family resemblances: Studies in the internal structure
  of categories. Cognitive Psychology  \textbf{7}(4),  573--605 (1975)

\bibitem{transfer}
Salman, H., Ilyas, A., Engstrom, L., Kapoor, A., Madry, A.: Do adversarially
  robust imagenet models transfer better? In: Adv. Neural Inform. Process.
  Syst. (NeurIPS) (2020)

\bibitem{certifiedsm}
Salman, H., Sun, M., Yang, G., Kapoor, A., Kolter, J.Z.: Denoised smoothing:
  {A} provable defense for pretrained classifiers. In: Adv. Neural Inform.
  Process. Syst. (NeurIPS) (2020)

\bibitem{carlinipart}
Sitawarin, C., Pongmala, K., Chen, Y., Carlini, N., Wagner, D.A.: Part-based
  models improve adversarial robustness. In: Int. Conf. Learn. Rep. (ICLR)
  (2023)

\bibitem{Apollocar3D}
Song, X., Wang, P., Zhou, D., Zhu, R., Guan, C., Dai, Y., Su, H., Li, H., Yang,
  R.: Apollocar3d: {A} large 3d car instance understanding benchmark for
  autonomous driving. In: IEEE Conf. Comput. Vis. Pattern Recog. (CVPR). pp.
  5452--5462 (2019)

\bibitem{vlpart}
Sun, P., Chen, S., Zhu, C., Xiao, F., Luo, P., Xie, S., Yan, Z.: Going denser
  with open-vocabulary part segmentation. arXiv preprint arXiv:2305.11173
  (2023)

\bibitem{labelsmoothing}
Szegedy, C., Vanhoucke, V., Ioffe, S., Shlens, J., Wojna, Z.: Rethinking the
  inception architecture for computer vision. In: IEEE Conf. Comput. Vis.
  Pattern Recog. (CVPR). pp. 2818--2826 (2016)

\bibitem{adv13}
Szegedy, C., Zaremba, W., Sutskever, I., Bruna, J., Erhan, D., Goodfellow,
  I.J., Fergus, R.: Intriguing properties of neural networks. In: Int. Conf.
  Learn. Rep. (ICLR) (2014)

\bibitem{adaptive20}
Tram{\`{e}}r, F., Carlini, N., Brendel, W., Madry, A.: On adaptive attacks to
  adversarial example defenses. In: Adv. Neural Inform. Process. Syst.
  (NeurIPS) (2020)

\bibitem{opc}
Tversky, B., Hemenway, K.: Objects, parts, and categories. Journal of
  Experimental Psychology: General  \textbf{113}(2), ~169 (1984)

\bibitem{in-sketch}
Wang, H., Ge, S., Lipton, Z.C., Xing, E.P.: Learning robust global
  representations by penalizing local predictive power. In: Adv. Neural Inform.
  Process. Syst. (NeurIPS). pp. 10506--10518 (2019)

\bibitem{wang2015unsupervised}
Wang, J., Zhang, Z., Xie, C., Premachandran, V., Yuille, A.: Unsupervised
  learning of object semantic parts from internal states of cnns by population
  encoding. arXiv preprint arXiv:1511.06855  (2015)

\bibitem{pang}
Wang, Z., Pang, T., Du, C., Lin, M., Liu, W., Yan, S.: Better diffusion models
  further improve adversarial training. In: Krause, A., Brunskill, E., Cho, K.,
  Engelhardt, B., Sabato, S., Scarlett, J. (eds.) Int. Conf. Mach. Learn.
  (ICML). vol.~202, pp. 36246--36263 (2023)

\bibitem{timm}
Wightman, R.: Pytorch image models.
  \url{https://github.com/rwightman/pytorch-image-models} (2019)

\bibitem{awp}
Wu, D., Xia, S., Wang, Y.: Adversarial weight perturbation helps robust
  generalization. In: Adv. Neural Inform. Process. Syst. (NeurIPS) (2020)

\bibitem{resnext}
Xie, S., Girshick, R.B., Doll{\'{a}}r, P., Tu, Z., He, K.: Aggregated residual
  transformations for deep neural networks. In: IEEE Conf. Comput. Vis. Pattern
  Recog. (CVPR). pp. 5987--5995 (2017)

\bibitem{wideresnet}
Zagoruyko, S., Komodakis, N.: Wide residual networks. In: Brit. Mach. Vis.
  Conf. (BMVC) (2016)

\bibitem{trades}
Zhang, H., Yu, Y., Jiao, J., Xing, E.P., Ghaoui, L.E., Jordan, M.I.:
  Theoretically principled trade-off between robustness and accuracy. In: Int.
  Conf. Mach. Learn. (ICML). vol.~97, pp. 7472--7482 (2019)

\bibitem{robustify}
Zhang, Y., Yao, Y., Jia, J., Yi, J., Hong, M., Chang, S., Liu, S.: How to
  robustify black-box {ML} models? {A} zeroth-order optimization perspective.
  In: Int. Conf. Learn. Rep. (ICLR) (2022)

\bibitem{MHP}
Zhao, J., Li, J., Cheng, Y., Sim, T., Yan, S., Feng, J.: Understanding humans
  in crowded scenes: Deep nested adversarial learning and {A} new benchmark for
  multi-human parsing. In: ACM Int. Conf. Multimedia (ACM MM). pp. 792--800
  (2018)

\bibitem{objparsing}
Zhao, Y., Li, J., Zhang, Y., Tian, Y.: Multi-class part parsing with joint
  boundary-semantic awareness. In: Int. Conf. Comput. Vis. (ICCV). pp.
  9176--9185 (2019)

\bibitem{ade20k}
Zhou, B., Zhao, H., Puig, X., Xiao, T., Fidler, S., Barriuso, A., Torralba, A.:
  Semantic understanding of scenes through the {ADE20K} dataset. Int. J.
  Comput. Vis. (IJCV)  \textbf{127}(3),  302--321 (2019)

\bibitem{humanadv}
Zhou, Z., Firestone, C.: Humans can decipher adversarial images. Nature
  Communications  \textbf{10}(1), ~1--9 (2019)

\end{thebibliography}

\clearpage
\setcounter{page}{1}
\appendix
\renewcommand{\thefigure}{S\arabic{figure}}
\renewcommand{\thetable}{S\arabic{table}}
\setcounter{table}{0} 
\setcounter{figure}{0} 


\title{PartImageNet++ Dataset: Scaling up Part-based Models for Robust Recognition \\ \texttt{Supplementary Material}} 

\author{Xiao Li\orcidlink{0000-0001-8992-4944} \and
Yining Liu \and
Na Dong \and
Sitian Qin \and
Xiaolin Hu\orcidlink{0000-0002-4907-7354}
}
\authorrunning{Li et al.}
\titlerunning{PartImageNet++ Dataset: Scaling up Part-based ...}
\institute{Tsinghua University, Beijing, China}

\maketitle

\section{Details on the Reused Annotations of PIN}
\label{sec:supp_reuse}

To reduce the annotation cost without compromising annotation quality, we selected and retained part annotations of some categories from the original PIN dataset \cite{partimagenet}, which originally consisted of 158 object categories. Our selection process involved several steps. First, we excluded all object categories in the PIN that belong to the super-categories \textit{snake}, \textit{car}, \textit{aeroplane}, and \textit{bottle}, since we found that the part annotations for these categories were insufficient for our purpose. For example, \textit{car} is annotated with \textit{side mirror}, \textit{body}, and \textit{tier} in PIN, whereas we required that \textit{window}, \textit{wheel}, \textit{front side}, \textit{left side}, \textit{right side}, \textit{back side}, and \textit{top side} should be annotated according to our annotation scheme. Second, for the remaining object categories belonging to the other super-categories, we carefully inspected their existing part annotations to see if they aligned with our specific annotation requirements. We further excluded categories like \textit{ram} where the existing annotations in PIN did not adequately capture important parts (\eg, \textit{horn} that we considered significant). Lastly, we removed categories in PIN with less than 100 annotated images to ensure a sufficient amount of annotations for each category.

After applying the above procedures, we retained part annotations for a total of 90 object categories, which formed part of the PIN++ dataset. Below is a list of the retained object categories in PIN++:

\noindent n01440764, n01443537, n01484850, n01491361, n01494475, n01608432, n01614925, n01630670, n01632458, n01641577, n01644373, n01644900, n01664065, n01665541, n01667114, n01667778, n01669191, n01685808, n01687978, n01688243, n01689811, n01692333, n01693334, n01694178, n01695060, n01697457, n01698640, n01855672, n02002724, n02009229, n02009912, n02017213, n02025239, n02058221, n02071294, n02085782, n02089867, n02090379, n02092339, n02096177, n02096585, n02097474, n02098105, n02099601, n02100583, n02101006, n02101388, n02102040, n02102973, n02109525, n02109961, n02112137, n02114367, n02120079, n02124075, n02125311, n02128385, n02129604, n02130308, n02132136, n02133161, n02134084, n02134418, n02356798, n02397096, n02480495, n02480855, n02481823, n02483362, n02483708, n02484975, n02486261, n02486410, n02487347, n02488702, n02489166, n02490219, n02492035, n02492660, n02493509, n02493793, n02494079, n02514041, n02536864, n02607072, n02655020, n02835271, n03792782, n04482393, n04483307.

For the excluded 68 object categories, we reannotated them to ensure improved quality and alignment with our annotation scheme. Refer to \cref{sec:supp_statistics} for a visual comparison between our new annotations and the original annotations in PIN for some categories.

\begin{table}[!t]
  \centering
  \caption{The annotation density of PIN++.}
  \small
 \setlength{\tabcolsep}{5pt}
  {
    \begin{tabular}{c|cccc}
    \toprule
     Number of part masks & 1-2 & 3-4 & 5-6 & 7+  \\
     \midrule
     Proportion ($\%$) & $46.27$ & $44.87$ & $8.66$ & $0.20$ \\
     \bottomrule
    \end{tabular}   
    }
  \label{tab:density}
\end{table}

\begin{figure*}[!t]
  \centering
   \includegraphics[width=0.98\linewidth]{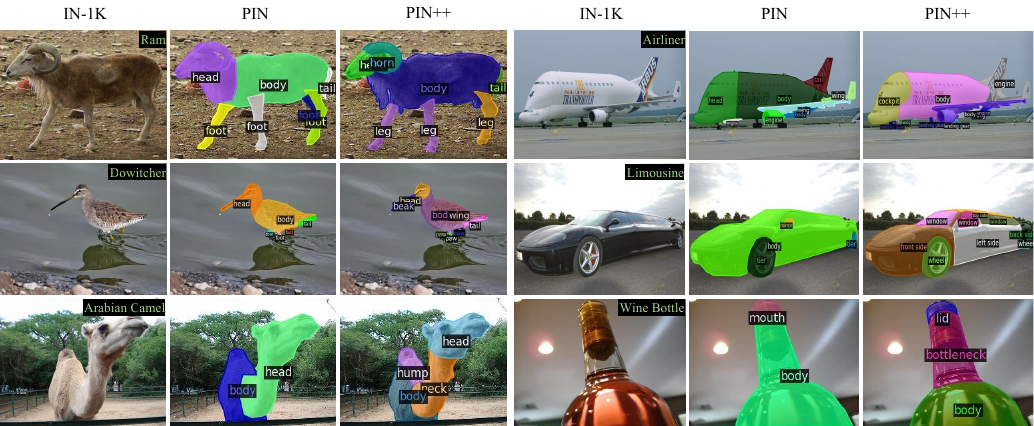}
   \caption{Visual comparison between annotations of PIN and PIN++. The object names are shown on the top-right of the IN-1K columns. The part names are shown in the PIN and PIN++ images.}
   \label{fig:compare}
\end{figure*}

\section{Additional Rules on Deciding Part Categories}
\label{sec:supp_part}

Besides utilizing the Wikidata and referring to the cognition of volunteers, as described in \cref{sec:annotationscheme}, we use additional rules to decide the part categories to annotate. Firstly, the combination of the part categories should form the complete object category. Secondly, the number of part categories is roughly set to be within the range of three to eight, except for special cases (\eg, \textit{flatworm}). When an object category is relatively simple and hard to annotate with more than three part categories, a certain part category can be added based on its function, shape, etc. For example, in the case of a \textit{maraca}, which consists of part categories \textit{head} and \textit{stick}, the \textit{stick} can be further divided into part categories \textit{joint} and \textit{handle} to better represent the gripping action for a \textit{maraca}. Thirdly, in cases where two part categories for an object overlap significantly, only the larger part category is annotated. For example, in the case of an \textit{acoustic guitar}, where the \textit{fingerboard} and \textit{strings} significantly overlap, we consider only the \textit{fingerboard} as a part category. With these rules, we can achieve high-quality part annotations for all categories in the IN-1K dataset while controlling the annotation cost.

\section{Additional Part Segmentation Principles}
\label{sec:supp_principle}

In addition to the three principles described in \cref{sec:annotationscheme}, the following principles were followed to ensure high-quality part segmentation annotations. Firstly, the inclusion relation of parts (\eg, \textit{horn} is included in \textit{head}) should be annotated using a dictionary format. This dictionary specifies the relationship between the smaller internal part category (key) and the larger part category (value). It enables researchers to understand the hierarchical structure of the annotated part categories. Secondly, if there are multiple objects of the target category within an image, all of them should be annotated while all other objects are treated as background elements. Finally, some images may need to be discarded due to their low quality, but these discarded images should be retained to verify whether the reason for discarding meets our requirements, so as to avoid excluding difficult examples.

The annotation results, including annotated masks, inclusion relations, and records of discarded images, were returned to inspectors in batches. If two out of ten randomly selected annotations for each category failed to meet the principles, the entire category’s annotations were rejected and re-annotated to ensure high-quality annotations. Annotations for one category may go through several iterations until they meet the requirements.

\begin{figure*}[!t]
  \centering
   \includegraphics[width=0.98\linewidth]{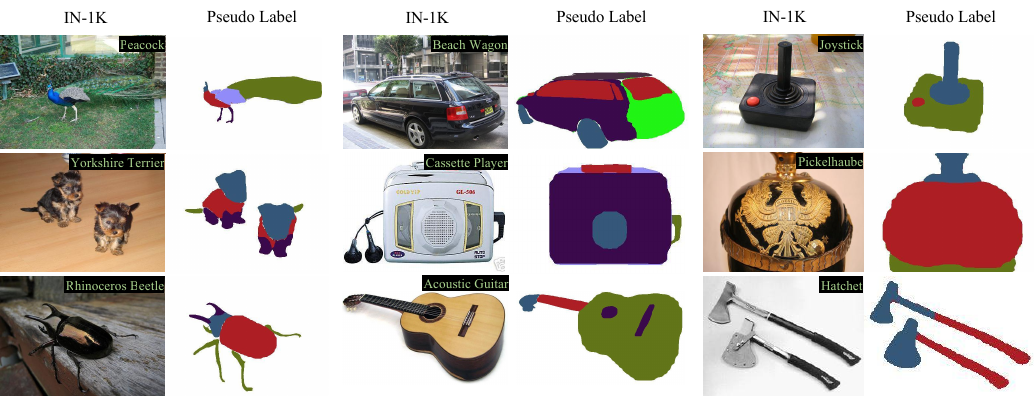}
   \caption{Visualization of pseudo part labels generated by a Mask R-CNN trained on PIN++. The object names are shown on the top-right of each image. The part names are hidden here for clarity.}
   \label{fig:pseudolabel}
\end{figure*}

\section{Additional Information about PIN++}
\label{sec:supp_statistics}

\noindent\textbf{Annotation cost and density.} 
PIN++ was annotated by 50 annotators who collectively invested approximately 8,000 hours in the annotation process. In addition, 10 volunteers were involved in determining the parts to be annotated and 5 inspectors were involved in the annotation quality inspection. During this process, 37,505 low-quality images were discarded, with each discarded image reviewed by two inspectors. With such efforts, PIN++ possesses a substantial number of part mask annotations, resulting in 406,364 mask annotations. We calculated the proportion of part mask annotations per image, which we called the density of annotations, as shown in \cref{tab:density}. Notably, over half of the images contain 3-6 part mask annotations. Given the extensive number of images in PIN++, it offers abundant training supervision for various part-related visual tasks.

\noindent\textbf{Inclusion relations.} 
PIN++ provides inclusion relations for annotated part masks that exhibit overlap. These relations are organized in a dictionary format, categorized by each object category, resulting in a total of 201 object categories with corresponding 317 inclusion relations. These inclusion relations serve as valuable resources for researchers to understand the hierarchical structure of the annotated part categories.

\noindent\textbf{Visual comparison between PIN and PIN++.} 
\cref{fig:compare} shows some visual annotations from the 68 object categories annotated in both PIN and PIN++ (see \cref{sec:supp_reuse}). The comparison indicates that PIN++ provides superior annotation quality compared to PIN. PIN++ exhibits finer-grained part segmentation and more accurate part names and effectively captures the distinctive characteristics of each object category.

\section{Visualization of Pseudo Part Labels}
\label{sec:supp_preudo}

\cref{fig:pseudolabel} illustrates the pseudo part labels generated by a Mask R-CNN trained on PIN++, as shown in \cref{pseudo-generation}. It indicates that supervised training on PIN++ yielded good part segmentation results on IN-1K and provides high-quality pseudo labels. Thus, we simply treat the pseudo labels and real part annotations equally when training MPM (see \cref{sec:modeldesign}). In addition, we provide quantitative results for pseudo-label quality in \cref{sec:ablation} and \cref{tab:pl_quality}.

\section{Training Recipe Details}
\label{sec:supp_recipe}
We improved the training recipe of ResNet-50-GELU to build a strong baseline. We implemented these using the \texttt{timm} \cite{timm} library. The details are as follows. 

\noindent\textbf{AT recipe}: The ResNet-50-GELU was optimized using stochastic gradient descent (SGD) with an initial learning rate of 0.2, a momentum of 0.9, and a cosine decay scheduler for the learning rate. The weight decay was set to be $1 \times 10^{-4}$. Data augmentation techniques, including random flipping and cropping, were applied during training. The model was trained for 80 epochs using 8 NVIDIA 3090 GPUs with a batch size of 512. The training process started with an initialization of a clean pretrained ResNet-50 checkpoint (on IN-1K) obtained from \texttt{torchvision}\footnote{\url{https://download.pytorch.org/models/resnet50-11ad3fa6.pth}}. Additionally, Exponential Moving Average (EMA) \cite{ema} with a decay of 0.9998 and label smoothing \cite{labelsmoothing} with a parameter of 0.1 were utilized. When training MPM, we used the same recipe for a fair comparison, except that we used an extra $L_{\rm{seg}}$ to introduce part-based supervision. Here $L_{\rm{seg}}$ used a Focal loss \cite{focal}, a variant of cross-entropy loss, to accelerate the convergence of part segmentation.

\noindent\textbf{Standard training recipe}: The standard training recipe closely resembles the AT recipe, except for the following differences. Instead of adversarial examples, clean examples were used during training. In addition, the models were trained for 100 epochs from scratch, without leveraging any pretrained checkpoints.

\section{Comparison Results on PIN and PIN++}
\label{sec:supp_comparison}

We conducted a comparison between the recognition accuracies of MPM trained on PIN++ and previous part-based models \cite{rock, carlinipart} trained on PIN. In line with previous works, AT for MPM was performed with an $l_{\infty}$ bound of $\epsilon = 8/255$. The training setup described in \cref{sec:setting} was followed, ensuring consistency across experiments. 

While MPM can perform classification on 1000 categories of PIN++, previous studies \cite{rock, carlinipart} reported results on subsets of PIN. To align with these studies, we masked out the other category channel of MPM and reported the results on the corresponding subsets. Specifically, the results on 125 categories of PIN were reported in \cref{tab:class125}, following the setting of Li \etal \cite{rock}. Additionally, the results on 11 super-categories of PIN were reported in \cref{tab:class11}, consistent with the setting of Sitawarin \etal \cite{carlinipart}. 

The results obtained with MPM trained on PIN++ demonstrate significant improvements over previous results achieved on PIN. However, we note that caution should be exercised when interpreting these results, as the data used for training are distinct and not directly comparable.

\begin{table}[!t]
  \centering
  \caption{Recognition accuracies (\%) of MPM and ROCK \cite{rock} on 125 categories of PIN. $l_\infty$ and $l_\infty (\times 2)$ indicates $l_\infty$ attacks with the bounds $\epsilon_\infty = 4/255$ and $\epsilon_\infty = 8/255$, respectively.}
  \small
 \setlength{\tabcolsep}{4pt}
  {
    \begin{tabular}{c|ccc}
     Method  &  Clean & $l_\infty$ & $l_\infty (\times 2)$  \\
     \midrule
     Li \etal \cite{rock} & $54.5$ &  $34.2$ & $17.3$ \\
     \hline
     ours (MPM) &   $\mathbf{70.0}$ & $\mathbf{42.3}$ & $\mathbf{18.9}$\\
     
    \end{tabular}   
    }
  \label{tab:class125}
\end{table}

\begin{table}[!t]
  \centering
  \caption{Recognition accuracies (\%) of MPM and the part-based model proposed by Sitawarin \etal \cite{carlinipart} on 11 super-categories of PIN.}
  \small
 \setlength{\tabcolsep}{4pt}
  {
    \begin{tabular}{c|ccc}
     Method  &  Clean &$l_\infty$ & $l_\infty (\times 2)$  \\
     \midrule
     Sitawarin \etal \cite{carlinipart} & $85.6$ & - & $39.4$ \\
     \hline
     ours (MPM) & $\mathbf{91.4}$ & $\mathbf{72.3}$ & $\mathbf{43.7}$ \\
     
    \end{tabular}   
    }
  \label{tab:class11}
\end{table}

\section{Robustness on OOD Datasets}
\label{sec:supp_datasets}

We introduce the OOD datasets used in this work first:
\begin{itemize}
    \item \textbf{ImageNet-A-Plus} (IN-A-Plus) \cite{in-a-plus}: This dataset consists of 3,286 images and serves as an improved version of ImageNet-A \cite{naturaladv}. IN-A-Plus comprises real-world and unmodified challenging images specifically curated for studying the robustness of classifiers to the internal variance of objects.
    \item \textbf{ImageNet-Sketch} (IN-Sketch) \cite{in-sketch}: IN-Sketch comprises 50,000 sketch-like images from the 1,000 categories in IN-1K. All images in this dataset are within a ``black and white'' color scheme.
    \item \textbf{Stylized-ImageNet} (SIN) \cite{sin}: SIN is a stylized variant of IN-1K, where different styles of artistic paintings are randomly applied through style transfer techniques to the original images. We utilized the validation set of SIN.
    
    \item \textit{\textbf{Image distortion}} (DIN) dataset \cite{modelvshuman}: DIN comprises 18,080 OOD images, covering 17 different OOD distortion scenarios including changes to image texture and various forms of synthetic additive noise.
\end{itemize}
These OOD datasets were selected to provide diverse and challenging scenarios for evaluating the OOD robustness.

\begin{table}[!t]
  \centering
  \caption{Comparison of recognition results between different models and human decisions \cite{modelvshuman}. The direction of the arrows indicates better alignment between the models and humans.}
  \small
  \setlength{\tabcolsep}{4pt}
  {
    \begin{tabular}{cc|ccc}
     Part & AT & Acc. Diff.$\downarrow$ & Obs. Consistency$\uparrow$ & Error Consistency$\uparrow$  \\
     \midrule
     $ $ & $ $ & $0.083$ & $0.665$ & $0.195$  \\
     $\checkmark$ & $ $ & $\mathbf{0.081}$ & $\mathbf{0.668}$ & $\mathbf{0.204}$ \\
     \hline
     $ $ & $\checkmark$ & $\textbf{0.069}$ & $0.676$ & $0.252$ \\
     $\checkmark$ & $\checkmark$ & $\textbf{0.069}$ & $\textbf{0.679}$ & $\textbf{0.261}$ \\
    \end{tabular}
    }
  \label{tab:modelvshuman}
\end{table}

\section{Evaluating Differences between Humans and Models}
\label{sec:supp_humanvsmodel}
Geirhos \etal \cite{modelvshuman} provided the human judgment data on several types of image distortions. We employed these data and their evaluation method, which involved comparing the models’ decisions on different distorted images with the actual judgments made by human observers. The results are shown in \cref{tab:modelvshuman}. Three metrics quantify how closely model predictions are aligned with the decisions of humans \cite{modelvshuman}: \textit{Accuracy Difference} measures the average accuracy disparity between a model and human observers across different image distortions. \textit{Observed Consistency} quantifies the agreement between model and human decisions on the same samples. It measures the fraction of samples where both the model and humans make the same decision, regardless of whether it is correct or not. \textit{Error Consistency} is the key metric, which measures the shared mistakes between the model and humans, indicating consistency in error patterns.  When the accuracies of both models and humans are high, they can achieve high observed consistency with very different decision strategies, while error consistency can still track whether there is above-chance consistency in decision-making. From the results under these metrics, we can conclude that our part-based models measurably improved alignment with the human recognition process, regardless of whether AT was used or not.

\section{Boosting Robustness on Downstream Tasks}
\label{sec:supp_downstream}

 We conducted similar experiments by using the checkpoints of the vanilla baseline and MPM to initialize the backbone (a ResNet-50) of a Faster R-CNN and subsequently performing downstream AT using the recipe proposed by Li \etal \cite{advod}. The results on the downstream object detection are shown in \cref{tab:downstream}. MPM enhances both the clean accuracy and adversarial robustness in object detection. Notably, there is a substantial boost in AP$_{50}$, a practical metric widely used in detection evaluations \cite{advod}. The results highlight the importance of investigating part-based models on the large-scale IN-1K again.

\begin{table}[!t]
  \centering
  \caption{Detection accuracies on MS-COCO \cite{coco}. $A_{\mathrm{cls}}$ and $A_{\mathrm{reg}}$ represent the attacks on the classification loss and regression loss of detectors, respectively. 
   }
  \small
 \setlength{\tabcolsep}{4pt}
  {
    \begin{tabular}{c|cc|cc|cc}
     \multirow{2}*{Initialization}  & \multicolumn{2}{c|}{Clean} & \multicolumn{2}{c|}{$A_{\mathrm{cls}}$}  & \multicolumn{2}{c}{$A_{\mathrm{reg}}$}  \\ 
     \cline{2-7}
             & AP  & \multicolumn{1}{c|}{AP$_{50}$}
             & AP  & \multicolumn{1}{c|}{AP$_{50}$}
             & AP  & \multicolumn{1}{c}{AP$_{50}$} \\
    \midrule
     \cite{advod} & $29.9$&$49.3$&$\mathbf{14.8}$&$25.5$&$19.7$&$40.5$\\
     \rowcolor{lightgray} ours (vanilla) &$29.9$&$49.8$&$14.5$&$25.4$&$19.3$&$40.2$\\
     \rowcolor{lightgray} ours (MPM) &$\mathbf{30.3}$&$\mathbf{50.3}$&$\mathbf{14.8}$&$\mathbf{25.9}$&$\mathbf{19.8}$&$\mathbf{40.7}$\\
    \end{tabular}   
    }
  \label{tab:downstream}
\end{table}

\begin{table}[!t]
  \centering
  \caption{Recognition accuracies (\%) of MPM on IN-1K trained with object segmentation mask and part segmentation mask.}
  \small

  \setlength{\tabcolsep}{4pt}
  {
    \begin{tabular}{c|cccc|c}
     Supervision & Clean & $l_\infty$ & $l_1$ & $l_2$ & Average  \\
     \midrule
     Object & $67.5$ & $37.3$ & $6.0$ & $23.6$ & $22.3$ \\
     Part & $\mathbf{67.8}$ & $\mathbf{39.1}$ & $\mathbf{6.2}$ &  $\mathbf{24.3}$ & $\mathbf{23.2}$ \\
    \end{tabular}
    }
  \label{tab:pl_ol}
\end{table}

\section{Additional Ablation Studies}
\label{sec:supp_others}

We introduce additional experiments on the part-based models.

\subsection{Part Label v.s. Object Label}
We were interested in whether part segmentation labels can be replaced by simpler object segmentation labels. Thus, we conducted a study to investigate this. Specifically, we replaced the supervision of part segmentation masks with the corresponding object segmentation masks while keeping the other components of the model unchanged. The results are shown in \cref{tab:pl_ol}. Comparing the model trained with part segmentation masks to the one trained with object segmentation masks, we observed that the latter achieved lower accuracies for both clean and adversarial images. These results emphasize the significance of more fine-grained part annotations rather than object segmentation annotations.

\begin{table}[!t]
  \centering
  \caption{Recognition accuracies (\%) of different methods on T-IN. Previous methods used WideResNet-28-10 with the input resolution of $64 \times 64$ while we used ResNeXt-50 with the input resolution of $224 \times 224$. FLOPs and the number of parameters during inference are listed. $l_\infty$ and $l_\infty (\times 2)$ indicates $l_\infty$ attacks with the bounds $\epsilon_\infty = 4/255$ and $\epsilon_\infty = 8/255$, respectively. The original results with label leakage are shown in \textcolor{newgray}{gray}.}
  \small
 \setlength{\tabcolsep}{4pt}
  {
    \begin{tabular}{c|cc|ccc}
     Method & Param. & FLOPs & Clean &$l_\infty$ & $l_\infty (\times 2)$  \\
     \midrule
     Gowal \etal \cite{generated} & \multirow{3}*{$36.6$M} & \multirow{3}*{$21.0$G} & \textcolor{newgray}{$61.0$} & \textcolor{newgray}{-} & \textcolor{newgray}{$26.7$} \\
     Wang \etal \cite{pang} &  & & \textcolor{newgray}{$65.2$} & \textcolor{newgray}{$48.3$} & \textcolor{newgray}{$31.3$} \\
     Wang \etal \cite{pang} &  & & $57.6$ & $38.4$ & $28.4$\\
     \hline
     ours (vanilla) & \multirow{2}*{$23.4$M} & \multirow{2}*{$4.3$G} & $67.9$ & $48.0$ & $27.6$\\
     ours (MPM) &  & & $\mathbf{69.0}$ & $\mathbf{48.9}$ & $\mathbf{28.5}$ \\
    \end{tabular}   
    }
  \label{tab:tiny_compare}
\end{table}

\begin{figure}[!t]
  \centering
   \includegraphics[width=0.6\linewidth]{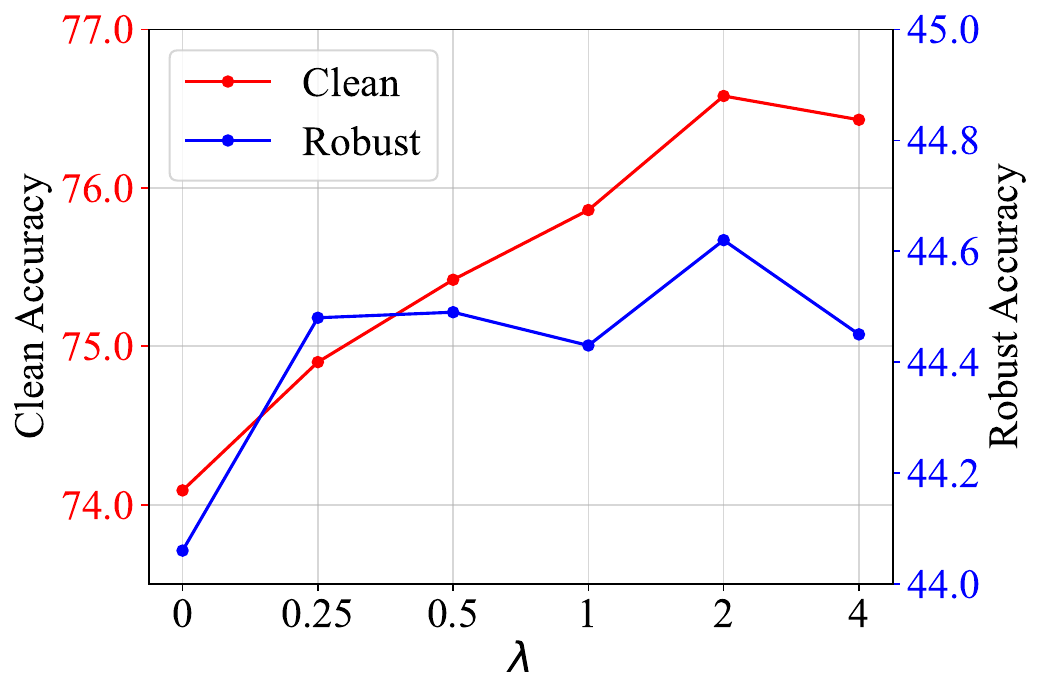}
   \caption{Recognition accuracies (\%) of MPM on T-IN with different $\lambda$. Here \textit{robust} is evaluated by the $l_\infty$ attack with a bound of $\epsilon = 4/255$.}
   \label{fig:tinyimagenet}
\end{figure}

\subsection{Comparisons on Tiny-ImageNet} 
We notice that most works on adversarial robustness \cite{trades, awp, generated, pang} are still evaluated only on small-scale, low-resolution datasets like CIFAR-10 \cite{cifar10}. Here we conducted additional experiments on another widely used low-resolution dataset Tiny-ImageNet (T-IN) \cite{tinyimagenet}. T-IN is a subset of IN-1K, consisting of 200 categories and 500 training images per category, all resized to $64\times64$ pixels. We created a similar dataset Tiny-PartImageNet++ (T-PIN++), with the released generation code\footnote{\url{https://github.com/jcjohnson/tiny-imagenet}} of T-IN. T-PIN++ includes part annotations (or pseudo part labels) for each training image and consists of the same categories and training images as T-IN, but without resizing, maintaining the original high-resolution of the IN-1K images.

We trained the vanilla ResNeXt-50 and MPM with the ResNeXt-50 \cite{resnext} as the backbone on T-PIN++ using the AT recipe described in \cref{sec:supp_recipe}. We then compared these models with SOTA methods on T-IN \cite{generated, pang}, which trained WideResNet-28-10 \cite{wideresnet} using additional 1M images generated by diffusion models \cite{diffusion}. The results are shown in \cref{tab:tiny_compare}. We noticed that the validation set of T-IN was derived from the training set of IN-1K, leading to label leakage in the evaluations of previous SOTA methods \cite{generated, pang}, as the diffusion models they used were trained on the entire IN-1K training set. Their original results are shown in gray in \cref{tab:tiny_compare}, and we reevaluated the results on IN-1K \texttt{val} set using the released checkpoint of Wang \etal \cite{pang} (Gowal \etal \cite{generated} were not reevaluated as we did not find the released checkpoint). We can see that MPM (with the input resolution of $224 \times 224$) surpasses previous SOTA methods (with the input resolution of $64 \times 64$ and significantly larger models) on T-IN on both adversarial robustness and clean accuracy. Furthermore, MPM achieves these superior results with about 1/5 FLOPs. These results highlight the significance of investigating adversarial robustness on high-resolution images, rather than relying solely on low-resolution datasets like CIFAR-10 or T-IN.

\subsection{Investigating the Influence of $\lambda$}
We investigated the effect of the hyper-parameter $\lambda$ (see the training objective described in \cref{sec:modeldesign}) on the performance of MPM. $\lambda = 0$ corresponds to the vanilla model without any part supervision and a larger $\lambda$ indicates a stronger emphasis on part supervision. Limited by computational resources, here the ablation study was performed with ResNet-50 on T-PIN++ (see above description on T-PIN++). The recognition accuracies of MPM with different $\lambda$ are shown in  \cref{fig:tinyimagenet}. We can observe that the clean accuracy of MPM increased with the increase of $\lambda$. Additionally, the adversarial robustness of MPM demonstrates low sensitivity to the specific value of $\lambda$ as long as $\lambda > 0$. In general, these results show the importance of part supervision.

\end{document}